\documentclass[10pt,twocolumn,letterpaper]{article}

\usepackage[pagenumbers]{cvpr} 

\definecolor{cvprblue}{rgb}{0.21,0.49,0.74}
\usepackage[pagebackref,breaklinks,colorlinks,allcolors=cvprblue]{hyperref}

\def\paperID{XXXXX} 
\def\confName{CVPR}
\def\confYear{2026}

\title{Claude Code-Driving Scenario Mining for the Argoverse 2 Challenge}

\author{Wei Deng, Caoshengzhe Xue, Shuaikun Liu, Zhaohong Liu, Mengshi Qi$^*$, Huadong Ma\\
State Key Laboratory of Networking and Switching Technology,\\
Beijing University of Posts and Telecommunications, China\\
{\tt\small \{dw-dengwei,xuecsz,liushuaikun1,liuzhaoh,qms,mhd\}@bupt.edu.cn}
}

\begin{document}
\maketitle
\begingroup
\renewcommand{\thefootnote}{}\footnotetext{$^*$Corresponding author.}
\endgroup
\begin{abstract}
We present our submission to the CVPR 2026 Argoverse 2 Scenario Mining Challenge. Our system uses a four-stage pipeline: (1) autonomous code generation via a \texttt{Claude Code} agent powered by \texttt{GLM~5.1}~\cite{glm5team2026glm5vibecodingagentic}, (2) iterative training set screening with Timestamp Balanced Accuracy threshold 0.8 to curate few-shot examples, (3) semantic code review by a separate \texttt{Claude Code} session, and (4) \texttt{Qwen3-VL}~\cite{bai2025qwen3vltechnicalreport} scene-level verification to filter false positives. We report results on the Argoverse 2 test set.
\end{abstract}

\section{Introduction}

The Argoverse 2 Scenario Mining Challenge is a CVPR 2026 competition tasking participants with localizing objects and temporal segments in autonomous driving logs given natural language descriptions. Built on the RefAV dataset~\cite{davidson2026refav}, which comprises 10,000 planning-centric queries spanning 1,000 driving logs from the Argoverse 2 Sensor Dataset, the competition features two tracks: \emph{Spatio-Temporal Localization} (metric: HOTA-Temporal) and \emph{Temporal Localization} (metric: Timestamp Balanced Accuracy).

We present a four-stage scenario mining pipeline. First, we use \texttt{Claude Code} as an autonomous coding agent powered by \texttt{GLM~5.1}~\cite{glm5team2026glm5vibecodingagentic} to generate scenario code from natural language prompts. Second, we iteratively screen generated code on the training set, retaining only prompt-code pairs that achieve Timestamp Balanced Accuracy above 0.8 (up to 5 iterations) as few-shot examples. Third, a separate \texttt{Claude Code} session reviews generated code for semantic correctness and false positive risks. Fourth, we introduce a \texttt{Qwen3-VL}~\cite{bai2025qwen3vltechnicalreport} verification stage that performs binary scene-level classification to filter false positives.

Our approach differs from existing work in two key aspects: (1) using an autonomous agent paradigm rather than single-pass LLM API calls, and (2) employing VLM-based verification as a post-execution filter that directly reduces false positive outputs.

\section{Method}

\begin{figure*}[t]
  \centering
  \includegraphics[width=\linewidth]{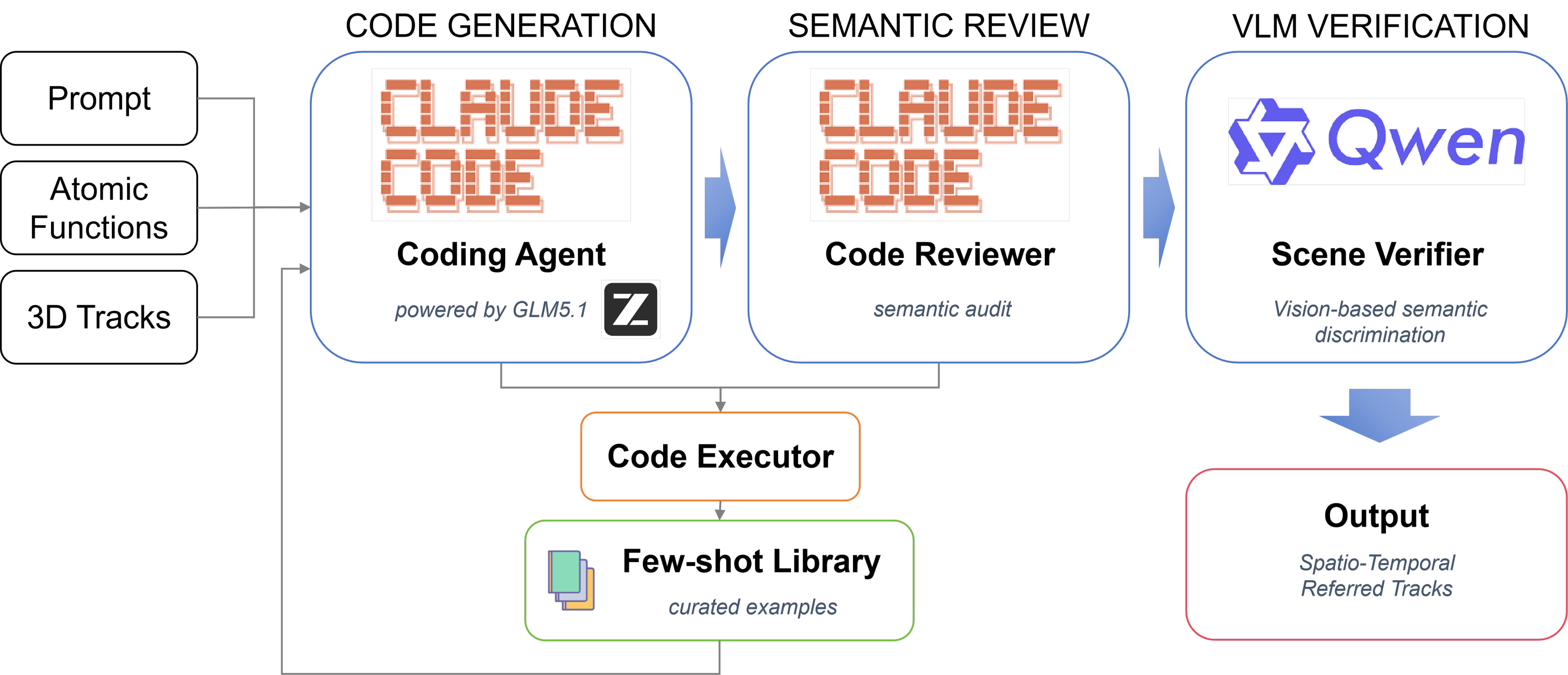}
  \caption{Overview of our framework.}
  \label{fig:overview}
\end{figure*}

We present a four-stage scenario mining pipeline built on the RefAV~\cite{davidson2026refav} atomic function framework. Our approach leverages an LLM-powered coding agent to autonomously generate scenario code, iteratively refines it using training set feedback, performs semantic code review, and validates outputs with a vision-language model to minimize false positives.

\subsection{Background: Composable Atomic Functions}

The RefAV framework provides a library of \emph{atomic functions} that describe spatial relations, kinematics, and map attributes. Each function is decorated with \texttt{@composable} or \texttt{@composable\_relational}, which handles parallelization over track candidates.

Spatial functions include \texttt{has\_objects\_in\_relative\_direction} (front, rear, left, right), \texttt{facing\_toward}, \texttt{heading\_toward}, \texttt{near\_objects}, and \texttt{being\_crossed\_by}. Kinematic functions include \texttt{turning}, \texttt{changing\_lanes}, \texttt{accelerating}, \texttt{has\_velocity}, and \texttt{stationary}. Map functions include \texttt{at\_stop\_sign}, \texttt{at\_pedestrian\_crossing}, \texttt{near\_intersection}, \texttt{on\_road}, and \texttt{in\_drivable\_area}. Logical combinators \texttt{scenario\_and}, \texttt{scenario\_or}, and \texttt{scenario\_not} allow Boolean composition.

\subsection{LLM Agent Code Generation}

Unlike prior work that uses LLM API calls to generate code in a single pass, we employ \texttt{Claude Code} as an autonomous coding agent powered by \texttt{GLM~5.1}~\cite{glm5team2026glm5vibecodingagentic}. The agent iteratively queries atomic function definitions using tool calls, analyzes the prompt requirements, and constructs the appropriate function composition. This agent-based approach allows the model to explore the function library, understand parameter semantics through tool-driven documentation lookup, and verify its own output before finalizing the code.

\subsection{Training Set Iterative Screening}

We perform iterative screening on the RefAV training set to build a high-quality few-shot example library. For each prompt, the system:

\begin{enumerate}
    \item Generates candidate code via the LLM agent.
    \item Executes the code on the training split and computes Timestamp Balanced Accuracy.
    \item If the score exceeds 0.8, the prompt-code pair is retained as a few-shot example.
    \item If not, the code is regenerated for up to 5 iterations before being discarded.
\end{enumerate}

This process produces a curated set of prompt-code pairs that serve as context examples in subsequent inference, improving code quality for unseen prompts.

\subsection{Semantic Code Review}

Before execution-based verification, a separate \texttt{Claude Code} session reviews each generated code snippet. The reviewer checks for semantic mismatches between the prompt and the code logic, identifies potential false positive sources (e.g., incorrect relationship direction, missing spatial constraints), and suggests or applies corrections. This two-tier agent design---generation followed by review---ensures both syntactic validity and semantic fidelity.

\subsection{VLM Scene-Level Verification}

Generated scenario outputs may still contain false positives when the LLM hallucinates object interactions. To address this, we introduce a verification stage using \texttt{Qwen3-VL}~\cite{bai2025qwen3vltechnicalreport}, a vision-language model. For each prompt-log pair, Qwen3-VL performs binary classification on the driving frames to determine whether the described event actually occurs.

If the VLM confirms the event, the code execution result is retained. Otherwise, the trajectory output is discarded and an empty prediction is produced. This effectively filters out false positive scenarios that pass the code execution stage but do not correspond to real events in the sensor data.

\section{Experiments}

\subsection{Experimental Setup}

\textbf{Dataset.} We evaluate on the Argoverse 2 Sensor Dataset test split using the RefAV~\cite{davidson2026refav} scenario mining annotations. The evaluation covers four metrics: HOTA-Temporal, HOTA-Track, Timestamp Balanced Accuracy (Timestamp BA), and Log Balanced Accuracy (Log BA).

\textbf{Code generation agent.} We use \texttt{Claude Code} as the coding agent with \texttt{GLM~5.1}~\cite{glm5team2026glm5vibecodingagentic} as the underlying language model. The agent performs tool-driven queries to explore the atomic function library before generating code.

\textbf{VLM verification.} \texttt{Qwen3-VL}~\cite{bai2025qwen3vltechnicalreport} performs binary scene classification on driving frames to determine whether the described event occurs.

\textbf{Trackers.} We obtain trajectories from the generated code using the Le3DE2E~\cite{wang2023technicalreportargoversechallenges} tracking model.

\subsection{Training Set Iterative Screening}

On the RefAV training set, we apply iterative code generation with up to 5 rounds per prompt. Code achieving Timestamp BA above 0.8 is retained as a few-shot example. This yields a small curated set of high-quality prompt-code pairs that serve as context in subsequent inference.

\subsection{Code Review Quality}

A separate \texttt{Claude Code} session reviews all generated code for semantic mismatches. The review identifies common error patterns including: incorrect relationship direction in relational functions, missing implicit map constraints, and unrealistic parameter thresholds. Corrected code replaces the original when issues are detected.

\subsection{Main Results}

Table~\ref{tab:leaderboard} shows the EvalAI test set leaderboard with all participating teams. Our team (MICTeam) achieves a HOTA-Temporal of 27.91, Timestamp BA of 69.65, and Log BA of 69.32. Our approach outperforms the official baselines (RefProg and SM-Agent) by a clear margin.

\begin{table}[t]
    \centering
    \caption{EvalAI test set leaderboard.}
    \label{tab:leaderboard}
    \resizebox{\columnwidth}{!}{%
    \begin{tabular}{cccccc}
        \hline
        Rank & Team & HOTA-Temp & HOTA-Trk & TS BA & Log BA \\
        \hline
        1 & HYU\_OASIS & 38.50 & 52.63 & 74.32 & 77.12 \\
        2 & MTL & 37.04 & 55.11 & 75.50 & 80.75 \\
        3 & MISI & 36.38 & 49.32 & 77.21 & 76.26 \\
        4 & Zelos\_Agent\_Driving & 33.45 & 45.23 & 72.45 & 73.66 \\
        5 & Lilly & 32.51 & 42.06 & 71.93 & 73.11 \\
        6 & DataClaw & 31.40 & 44.40 & 74.05 & 79.10 \\
        7 & Predicate & 31.09 & 40.85 & 72.25 & 70.17 \\
        8 & blackTea & 30.87 & 41.98 & 72.39 & 75.23 \\
        9 & lazydogs & 30.79 & 42.12 & 71.44 & 73.73 \\
        10 & badr & 30.62 & 41.17 & 70.88 & 71.93 \\
        \textbf{11} & \textbf{MICTeam (Ours)} & \textbf{27.91} & \textbf{37.88} & \textbf{69.65} & \textbf{69.32} \\
        12 & RefProg & 26.27 & 36.18 & 68.07 & 70.46 \\
        13 & SM-Agent & 23.25 & 31.15 & 66.95 & 67.66 \\
        14 & SM-Fusion & 18.79 & 25.86 & 63.51 & 65.21 \\
        \hline
    \end{tabular}%
    }
\end{table}

\section{Conclusion}

We presented a Claude-Code-driven scenario mining system for the CVPR 2026 Argoverse 2 Scenario Mining Challenge. Our approach combines composable atomic functions with automated code generation, validated few-shot examples, scene-level vision-based verification, and iterative refinement. We achieved 27.91 scores in the HOTA-Temp metric.

\nocite{RDCLR,DC-SAM,yttip,HP-MCoRe,Qi_2026_CVPR,yun2024weakly,liao2026improving,8733019,9351755,10.1145/3394171.3416269,safedriverag,Lv_2025_CVPR}
{
    \small
    \bibliographystyle{ieeenat_fullname}
    \bibliography{main}
}


\end{document}